\documentclass[letterpaper, 10 pt, conference]{ieeeconf}

\IEEEoverridecommandlockouts
\usepackage{xcolor}
\def\BibTeX{{\rm B\kern-.05em{\sc i\kern-.025em b}\kern-.08em
    T\kern-.1667em\lower.7ex\hbox{E}\kern-.125emX}}

\makeatletter

\let\proof\@undefined
\let\endproof\@undefined
\makeatother

\usepackage[sort,compress]{cite}

\usepackage{optidef} % For optimization problem writing

\usepackage{amsmath,mathrsfs} % assumes amsmath package installed
\usepackage{amssymb}  % assumes amsmath package installed
\usepackage{bbm}
\usepackage{booktabs}
\usepackage[ruled]{algorithm2e}
\usepackage[noend]{algorithmic}
\usepackage{breqn}
\usepackage{diagbox}
\usepackage[hidelinks]{hyperref}
\usepackage[normalem]{ulem} % Ulem package to get strike through font
\def\vd{3pt}

\usepackage{amsthm}
\theoremstyle{definition}
\newtheorem{definition}{Definition}%[section]
\newtheorem{theorem}{Theorem}%[section]
\newtheorem{lemma}[]{Lemma}

\mathchardef\mhyphen="2D % Define a "math hyphen"

\usepackage{booktabs}%,caption}
\usepackage[flushleft]{threeparttable}

\usepackage{scalerel,stackengine}

\definecolor{mypink}{rgb}{0.858, 0.188, 0.478}
\definecolor{mygreen}{rgb}{0.2, 0.7, 0.2}
\definecolor{mycol}{rgb}{0.8, 0.5, 0.5}
\definecolor{mygray}{rgb}{0.5, 0.5, 0.5}

\newcommand{\nar}[1]{\scalebox{.9}[1.0]{\texttt{#1}}} % Narrow texttt text

\newcolumntype{C}{>{$}c<{$}} % math-mode version of "c" column type

\begin{document}

\title{%Optimal Robot Task Planning with Bounded Waiting \\ and Limited Human Assistance
Optimal Robot Path Planning In a Collaborative Human-Robot Team \\ with Intermittent Human Availability
%
% Optimal Robot Task Planning with Limited Human Assistance
%
% Optimal Planning of Robot Tasks
%
% Robot Task Planning
}
\author{Abhinav Dahiya and Stephen L.\ Smith
\thanks{This research is supported in part by the Natural Sciences and Engineering Research Council of Canada (NSERC)} 
\thanks{Abhinav Dahiya and Stephen L. Smith are with Department of Electrical and Computer Engineering, University of Waterloo, Waterloo
(\href{mailto:abhinav.dahiya@uwaterloo.ca}{abhinav.dahiya@uwaterloo.ca}, \href{mailto:stephen.smith@uwaterloo.ca}{stephen.smith@uwaterloo.ca})}%
}

\maketitle

\begin{abstract}
This paper presents a solution for the problem of optimal planning for a robot in a collaborative human-robot team, where the human supervisor is intermittently available to assist the robot in completing tasks more quickly.  Specifically, we address the challenge of computing the fastest path between two configurations in an environment with time constraints on how long the robot can wait for assistance.  To solve this problem, we propose a novel approach that utilizes the concepts of budget and critical departure times, which enables us to obtain optimal solution while scaling to larger problem instances than existing methods.   We demonstrate the effectiveness of our approach by comparing it with several baseline algorithms on a city road network and analyzing the quality of the obtained solutions.  Our work contributes to the field of robot planning by addressing a critical issue of incorporating human assistance and environmental restrictions, which has significant implications for real-world applications.

% In this paper, we present the problem of robot planning in a time-dependent graph, where edges can be traversed either autonomously or with the help of a remote operator. We model the availability of the operator as an arbitrary function of time, and an edge can only be traversed with operator assistance if the operator is available for the entire duration of that edge. Each edge is characterized by different robot speeds under autonomous and assisted operation. Waiting is allowed at vertices, bounded by a maximum waiting duration defined for each vertex. To solve this problem, we first provide a greedy graph search method as a quick and simple solution, and then use the notion of budget and critical departure times to obtain the optimal solution while still being decently fast.  We present the performance and discuss the quality of solutions provided by the two methods using an example of a robot navigating through a city road network.  We also discuss how this problem can be solved using existing methods with slight modifications, and demonstrate the effectiveness of our approach in large graphs.
\end{abstract}

% \begin{IEEEkeywords}
% Human-robot collaboration, task planning, time-dependent graphs, limited human assistance.
% % Operator allocation, multi-robot supervision, restless bandits, Whittle index
% \end{IEEEkeywords}

\section{Introduction}
\begin{figure}
    \centering
    \includegraphics[width=0.825\columnwidth]{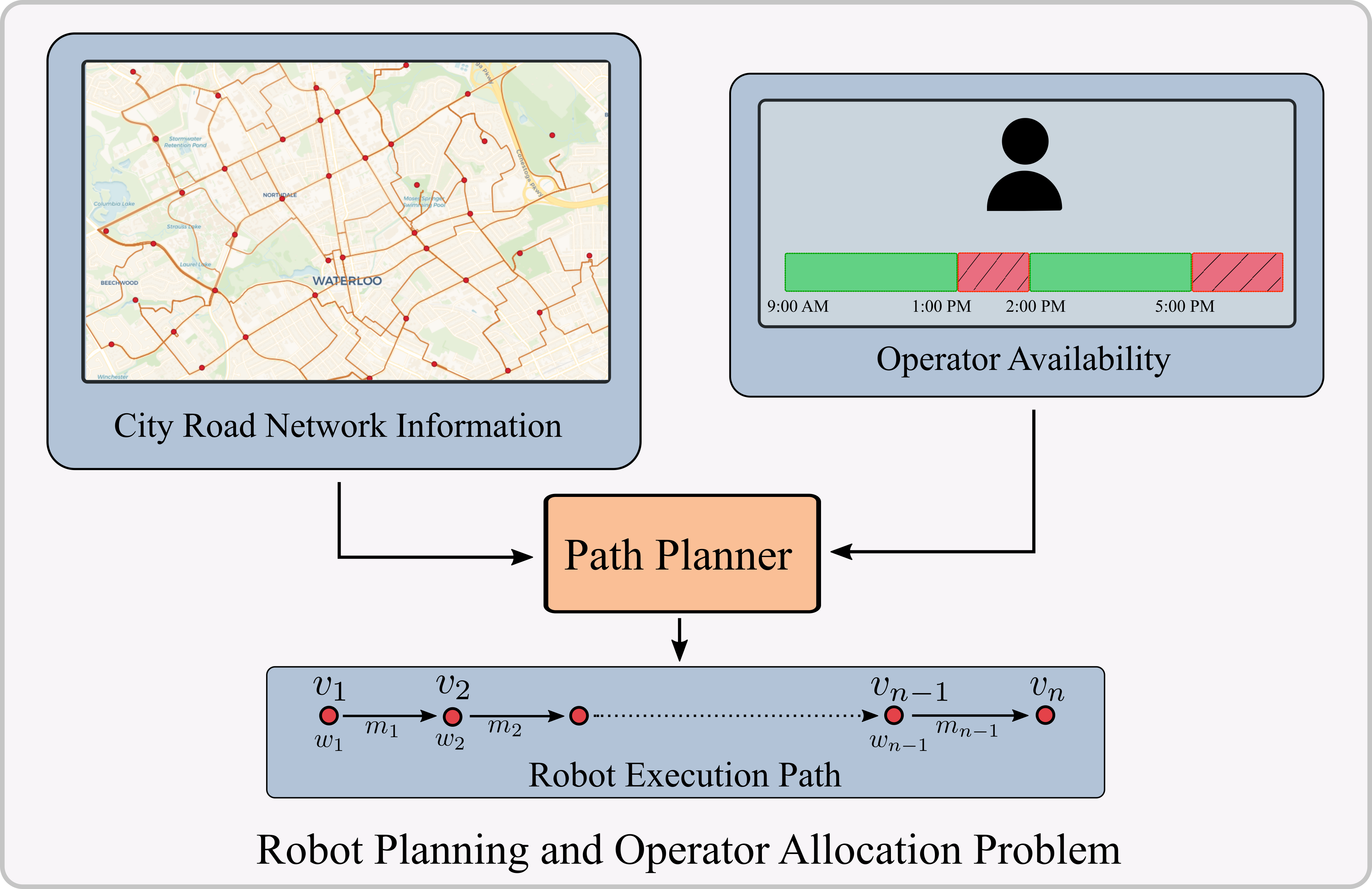}
    \caption[Problem Overview]{Problem Overview: Given the information of a city road network\footnotemark ~with speed and waiting restrictions, and the availability of human assistance, the objective is to determine a path that results in the fastest arrival at a goal location. The path consists of three components: (1) the vertices or locations to navigate through, denoted as $(v_1, v_2, \ldots, v_n)$; (2) the amount of waiting required at intermediate locations, denoted as $(w_1, w_2, \ldots, w_{n-1})$; and (3) whether to use human assistance, denoted as $(m_1, m_2, \ldots, m_{n-1})$.  In this paper, we propose a novel and efficient planning algorithm to obtain the required path.}
    \label{fig:front_fig}
\end{figure}

% In the last two decades, robotics has seen significant advancements in autonomy, with applications ranging from industrial and urban spaces to social settings \cite{royakkers2015literature, mintrom2022robots, dahiya2023survey}.  
% However, a significant challenge remains in enabling robots to navigate in dynamic environments effectively and safely.  While robots can operate autonomously, human assistance may still be required to enhance safety, efficiency, or to comply with regulations.  
Robots have come a long way in the past decades, with increasing levels of autonomy transforming the way they operate in various domains, from factories and warehouses to homes and public spaces \cite{royakkers2015literature, mintrom2022robots, dahiya2023survey}.  However, navigating dynamic environments effectively continues to be a formidable challenge.  Despite the significant strides made in robot autonomy, human oversight remains vital in enhancing safety, efficiency or to comply with regulatory requirements.
For example, a robot navigating through an urban environment must abide by traffic regulations and may require human assistance in busy or construction areas to ensure safety or expedite operations. Similarly, in an exploration task, robots may require replanning due to changes in the environment, while the supervisor has already committed to a supervision schedule for other robots and is only intermittently available.  By considering the operator's availability and environmental restrictions, robots can plan their paths more efficiently, avoid unnecessary waiting and decide when to use human assistance.  Figure~\ref{fig:front_fig} shows the problem overview with an example of a robot navigating in a city.  However, the presented problem can be generalized to any arbitrary task which can be completed via different sub-tasks defined using precedence and temporal constraints.

% \begin{figure}
%     \centering
%     \includegraphics[width=0.8\columnwidth]{Figures/DrawnMap.png}
%     \caption{A robot navigating in an urban environment may encounter different road features and restrictions in the form of speed limits, waiting/stopping rules and road closures.  This figure shows the road network of the city of Waterloo along with school zones (marked orange), commercial zones (marked purple) and construction zones (marked yellow).  Three possible paths between location $A$ and location $B$ are shown in the figure (in blue, black and red), where solid lines denote robot's autonomous operation while dashed lines denote assisted operation.  Depending on the availability of a remote supervisor, the optimal (fastest) path between two locations can change.  The map is generated using \href{https://www.mapbox.com/about/maps/}{Mapbox} and \href{http://www.openstreetmap.org/about/}{OpenStreetMap}. Road closure data is obtained from the Region of Waterloo website. \todo{Add mapbox logo in the image.}}
%     \label{fig:drawnMap}
% \end{figure}
We consider the problem of robot planning with the objective of finding the fastest path between two configurations. We demonstrate our approach through an example of robot navigation in an urban environment with intermittent operator availability, varying travel speeds, and waiting limits.  
Specifically, we consider a city road network where the robot can traverse through different locations either autonomously or with the assistance of a human supervisor, each taking different amounts of time. However, the supervisor is only available at certain times, and the robot has a limited amount of time to wait at a location before it must move on to its next destination.
By formulating the problem in this way, we aim to address the challenge of collaborative robot planning in real-world environments where the availability of human supervisors may be limited and thus can affect the optimal route for the robot.  In this paper, we present a method to compute the fastest path from one location to another while accounting for all these constraints\footnotetext{The map shown in Fig.~\ref{fig:front_fig} shows the road network of the city of Waterloo, generated using \href{https://www.qgis.org/en/site/index.html}{QGIS}, \href{https://www.openstreetmap.org/about/}{OpenStreetMap} and \href{https://openrouteservice.org/}{OpenRouteService}.}.

The problem of robot path planning with operator allocation in dynamic networks is inspired by real-world scenarios where the availability of human assistance and thus the robot's speed of travel and its ability to traverse certain paths can change over time, e.g., \cite{riley2021assessment}.  Traditional methods, such as time-dependent adaptations of the Dijkstra's algorithm, are not designed to handle situations in dynamic environments where waiting is limited, and the task durations may not follow the first-in-first-out (FIFO) property \cite{wang2019time}.  This means that a robot may arrive at its target location earlier by departing later from its previous location, for example, by using human assistance.  
To address these challenges, we draw on techniques from the time-dependent shortest path literature to solve the problem. Unfortunately, existing optimal solution techniques are severely limited by their computational runtime. In this paper, we propose a novel algorithm that is guaranteed to find optimal solution and runs orders of magnitude faster than existing solution techniques.

% Given the limitations of existing approaches, we propose a novel graph search algorithm to solve our problem of robot planning under the constraints of bounded waiting and intermittent assistance availability.

\emph{Contributions:} Our main contributions are as follows:

1) We propose a novel graph search algorithm for the collaborative planning problem with intermittent human availability.  The algorithm operates by intelligently selecting the times of exploration and by combining ranges of arrival times into a single search node.
% We introduce the concept of \textit{critical departure times} to enable efficient exploration during the graph search while ensuring an optimal arrival at the goal.

2) We provide the proof that the algorithm generates optimal solutions.

% We leverage the concept of \textit{budget} to capture a range of possible arrival times at a vertex using a single node, which leads to a significant reduction in redundant nodes and efficient search.

3) We demonstrate the effectiveness of our approach in a city road-network, and show that it outperforms existing approaches in terms of computational time and/or solution quality.

\section{Background and Related Work}
In this section, we discuss some relevant studies from the existing literature in the area of robot planning with human supervision/collaboration.  We also look into how the presented problem can be solved using existing techniques from related fields.  
% \todo{Add these: Safety Concerns Emerging from Robots Navigating in Crowded Pedestrian Areas;

% This requires the robots to navigate a number of challenges, including maintaining a balance between speed and safety, obeying traffic regulations, and dealing with dynamic and unpredictable environments \cite{calo2016robotLaw}.  Human supervision in such environments can enable faster speeds of operation or is required to help ensure a desired level of safety.  Even though completely autonomous systems are not currently ready for deployment in human-shared environments \cite{drew2021multi}, research is being carried out so that human involvement is not required for all robots all the time.  This requires identifying when to allocate human supervision to a particular robot/autonomous vehicle.  In the literature, we find studies that aim to do this based on recorded data \cite{swamy2020scaled}, transition probabilities \cite{dahiya2022scalable} or reachability analysis \cite{hickert2021cooperation}.  Past studies have also looked into scheduling of human assistance among multiple robots working on a sequence of tasks \cite{cai2022scheduling}.

\emph{Planning with Human Collaboration:} The problem of task allocation and path planning for robots operating in collaboration with humans has been studied extensively in recent years. Researchers have proposed various approaches, such as a data-driven approach for human-robot interaction modelling that identifies the moments when human intervention is needed \cite{swamy2020scaled}, and a probabilistic framework that develops a decision support system for the human supervisors, taking into account the uncertainty in the environment \cite{dahiya2022scalable}. In the context of autonomous vehicles, studies have investigated cooperative merging of vehicles at highway ramps \cite{hickert2021cooperation} and proposed a scheduling algorithm for multiple robots that jointly optimize task assignments and human supervision \cite{cai2022scheduling}. 

Task allocation is a common challenge in mixed human-robot teams across various applications, including manufacturing \cite{fusaro2021integrated}, routing \cite{hari2020approximation}, surveying \cite{mau2007scheduling}, and subterranean exploration \cite{riley2021assessment}.  In addition, the problem of computing the optimal path for a robot under time-varying human assistance bears similarity to queuing theory applications, such as optimal fidelity selection \cite{gupta2019optimal} and supervisory control of robots via a multi-server queue \cite{powel2012multiserver}.  
These studies provide insights into allocating assistance and path planning for robots in collaborative settings, but do not address our specific problem of computing the optimal path for a robot under bounded waiting and intermittent assistance availability. Additionally, our problem differs in that the robot can operate autonomously even when assistance is available, i.e., the collaboration is optional.

% The presented problem also bears similarity with several problems in queuing theory applications such as optimal fidelity selection where a human operator services a queue of tasks can choose between different levels of servicing quality \cite{gupta2019optimal}, and supervisory control of robots by human operators via a multi-server queue \cite{powel2012multiserver}. 

% This assurance of safety can not only be applicable to road navigation but can also be helpful in other (semi)autonomous systems in warehouse or healthcare setting.

\emph{Time-Dependent Shortest Paths:} 
The presented problem is also related to time-dependent shortest path (TDSP) problems, which aim to find the minimum cost or minimum length paths in a graph with time-varying edge durations \cite{wang2019time, dean2004algorithms}.  Existing solution approaches include planning in graphs with time-activated edges \cite{wang2019time}, implementing modified $A^*$ \cite{zhao2008algorithm}, and finding shortest paths under different waiting restrictions \cite{orda1990shortest, ding2008finding}.  Other studies have explored related problems such as computing optimal temporal walks under waiting constraints \cite{bentert2020efficient}, and minimizing path travel time with penalties or limits on waiting \cite{he2021time}.  
% There are also several studies that discuss the complexity of time-dependent shortest path problems under different settings \cite{casteigts2015expressivity, foschini2011complexity}. 
Many studies in TDSP literature have addressed the first-in-first-out (FIFO) graphs \cite{wang2019time}, while others have explored waiting times in either completely restricted or unrestricted settings.  However, the complexities arising from non-FIFO properties, bounded waiting and the need to make decisions on the mode of operation, i.e., autonomous or assisted, have not been fully addressed in the existing literature \cite{he2021time, foschini2011complexity, ding2008finding}.
% While these approaches provide valuable insights, they do not address our specific problem, which involves autonomous and assisted traversal of each edge and accounts for the time spent waiting when calculating the path length.  

% However, the problem presented in this paper has some unique features.  First, the time-dependency in our problem results from the change in operator availability instead of the edge travel times arbitrarily changing.  Second, we have a choice in the mode of operation to use.  This means even if the operator is available we may choose to run the robot autonomously.  
The most relevant solution technique that can be used to solve our problem is presented in \cite{cai1997time}, which solves a TDSP problem where the objective is to minimize the path cost constrained by the maximum arrival time at the goal vertex.  This method iteratively computes the minimum cost for all vertices for increasing time constraint value.  A time-expanded graph search method \cite{ford1958constructing} is another way of solving the presented problem by creating separate edges for autonomous and assisted modes.  We discuss these two methods in more detail in Sec.~\ref{sec:sim}.  As we will see, the applicability of these solution techniques to our problem is limited due to their poor scalability for large time horizons and increasing graph size.

\section{Problem Definition}
% There's a directed graph $G = (V,E)$. Each edge $e\in E$ has two travel times, an autonomous one $\tau(e,0)$ and a supervised one $\tau(e,1)$. We assume that $\tau(e,0) \geq \tau(e,1)$. Each vertex $v$ also has a maximum allowed waiting time $\overline{w}_v$. 

% There's a global operator availability function $\mu$ which tells if an operator is available at a given time or not. {The operator is considered to be available to traverse an edge if it is available to traverse the edge completely.  Whenever a robot is traversing an edge, if the operator is available for that edge, the robot takes the supervised travel time to traverse the edge, and if the operator is not available it takes the autonomous travel time.}\footnote{We can think of having different implementations of the availability function, e.g., as a time to binary function: $\real_{>=0} \rightarrow \{0,1\}$, or as a time to next operator availability change time along with current availability: $\real_{\geq 0} \rightarrow \real_{\geq 0} \times \{0,1\}$}. 
% Based on this global operator availability function and supervised edge travel time, we can find the operator availability for an edge $e$ for a given departure time $t^D$ as $\mu(e,t^D)$.

The problem can be defined as follows.  We are given a directed graph $G = (V,E)$, modelling the robot environment, where each edge $e \in E$ has two travel times corresponding to the two modes of operation: an autonomous time $\tau(e,0)$ and an assisted time $\tau(e,1)$, with the assumption that $\tau(e,0) \geq \tau(e,1)$.  When starting to traverse an edge, the robot must select the mode of operation for the traversal that is used for the entire duration of the edge.  While the autonomous mode is always available, the assisted mode can only be selected if the supervisor is available for the entire duration of the edge (under assisted mode).  The supervisor's availability is represented by a binary function $\mu$, with $\mu([t_1,t_2])=1$ indicating availability during the time window $[t_1,t_2]$, and 0 otherwise. %The availability of the supervisor is modeled as a known binary-valued function $\mu$, where for a given time window $[t_1,t_2]$, $\mu([t_1,t_2])=1$ if the operator is available during the entire window, and otherwise.
Additionally, at each vertex $v \in V$, the robot can wait for a maximum duration of $\overline{w}_v \geq 0$ before starting to traverse an outgoing edge.  %Note that, here we are considering only integer times. However the unit of time is arbitrary.

% The availability of a remote operator is also a factor in this problem, which is modeled by the availability function $\mu : \mathbb{Z}_{\geq0} \rightarrow \{0,1\}$. This function indicates whether the operator is available at a given time or not. If the operator is available $(\mu=1)$, a robot can traverse an edge in the supervised mode, which has a shorter travel time than the autonomous mode. If the operator is not available $(\mu=0)$, the robot must use the autonomous mode to traverse the edge. For an edge to be traversed under supervision, the operator must be available for the entire duration of the edge.
% The availability of the operator for a specific edge $e$ at a given departure time $t^D$ can be found using the function $\mu(e,t^D)$.

The robot's objective is to determine how to travel from a start vertex to a goal vertex. 
This can be represented as an \textit{execution path} $\mathcal{P}$, specified as a list of edges to traverse, the amount of waiting required at intermediate vertices and the mode of operation selected for each edge. % the corresponding waiting times at the intermediate vertices and the mode of operation chosen at each edge.  
The objective of this problem is to find an execution path (or simply path) from a start vertex $s \in V$ to a goal vertex $g \in V\setminus\{s\}$, such that the arrival time at $g$ is minimized.
% Note that the waiting time and mode of operation are not relevant for the goal vertex, and a valid execution path must satisfy certain conditions.  
%The problem objective is given below:
% \begin{equation}
%     \min \sum_{i=1}^{k-1}  t_w(v_i) + t(v_i, v_{i+1}),
% \end{equation}
% where $v_1 = s, v_k = g$, $t_w(v_i)$ is the actual waiting done at vertex $v_i$ and $t(v_i, v_{i+1})$ is the actual travel time from $v_i$ to $v_{i+1}$.
% % taking into account the waiting time at the beginning of each edge and ensuring that each waiting time does not exceed the maximum allowed waiting time.  
% Given a set $\hat{\mathcal{P}}$ of possible paths $\mathcal{P} \coloneqq \langle(v_1,t_1,w_1,m_1), (v_2,t_2,w_2,m_2), \ldots,$ $(v_n,t_n,w_n,m_n)\rangle$ of arbitrary length $n$, we can write the problem objective as follows: %from the set of all valid execution paths $\hat{\mathcal{P}}$, and denoting the arrival time at a vertex $v$ under path $\mathcal{P}$ as $t^A_v(\mathcal{P})$,

Given a set $\hat{\mathcal{P}}$ of all possible paths $\mathcal{P}$ of arbitrary length $n$, such that $\mathcal{P} \coloneqq \langle(v_1,t_1,w_1,m_1), (v_2,t_2,w_2,m_2), \ldots,$ $(v_n,t_n,w_n,m_n)\rangle$, we can write the problem objective as follows: 

% Given a set $\hat{\mathcal{P}}$ of all possible ordered lists of tuples $\langle(v_1,t_1,w_1,m_1), (v_2,t_2,w_2,m_2), \ldots,$ $(v_n,t_n,w_n,m_n)\rangle$ of arbitrary length $n$, we can write the problem objective as follows:
\begin{align*}
    \min_{\mathcal{P} \in \hat{\mathcal{P}}} \;\;\;\quad &\; t_n \\[\vd] %\coloneqq \sum_{i=1}^{n-1}w_i + \tau(e_{v_i v_{i+1}}, m_i). \nonumber\\
    s.t. \quad v_1 &= s, \; v_n = g\\[\vd]
    e_{v_i v_{i+1}} &\in E  \;\,\qquad\quad\forall~i\in[1, n-1]\\[\vd]
    % t^D_i &= t^A_i + w_i \quad\forall~i\in\mathcal{P}\\[\vd]
    t_{i+1} &= t_i + w_i + \tau(e_{v_i v_{i+1}}, m_i) \quad\forall~i\in[1, n-1]\\[\vd]
    w_i &\leq \overline{w}_{v_i} \;\,\,\quad\quad\forall~i\in[1,n-1]\\[\vd]
    m_i = 1 &\Rightarrow \mu([t_i + w_i, t_{i+1}])=1 \;\;\;\quad\forall~i\in[1,n-1].
    % m_i = 1 &\Rightarrow \mu(t) = 1~\forall~t \in [t_i + w_i, t_{i+1}]~\forall~i\in[1,n-1].
\end{align*}
The first constraint ensures that the path starts at $s$ and ends at $g$.  The second constraint ensures that the topological path is valid in the graph.  The third constraint ensures that the path does not violate travel duration requirements at any edge. Fourth constraint ensures that the waiting restrictions are met at each vertex. Finally, the fifth condition ensures that an edge can only be assisted if the operator is available at least until the next vertex is reached.
% Or maybe we can write it as:
% \begin{align*}
%     \arg\min_{\mathcal{P} \in \hat{\mathcal{P}}} \,\,t^A_g(\mathcal{P}) \coloneqq \sum_{i=1}^{n-1} t_i + w_i + \tau(e_{v_i v_{i+1}}, m_i).
% \end{align*}
% 
% Denoting arrival time at a vertex $v$ under path $\mathcal{P}$ as $t^A_v(\mathcal{P)}$, and the set of all valid execution paths $\hat{\mathcal{P}}$, we can write the problem objective as follows:
% \begin{align*}
%     \arg\min_{\mathcal{P} \in \hat{\mathcal{P}}} \,\,t^A_g(\mathcal{P)}.
% \end{align*}
% Or maybe we can write it as:
% \begin{align*}
%     \text{Minimize:}\qquad & t^A_g(\mathcal{P)} \nonumber\\
%     \text{Subject to:}\quad\;
%     \mathcal{P} &\in \hat{\mathcal{P}}\\[\vd]
%     % \mathcal{P} &= \Big\langle(v_1,t_1, w_1, m_1), \\ &\qquad\quad (v_2,t_2, w_2, m_2), \ldots, (v_n,t_n, w_n, m_n)\Big\rangle\\[\vd]
%     \mathcal{P}.v_1 &= s\\[\vd]
%     \mathcal{P}.t_1 &= 0.
%     % v_n &= g\\[\vd]
%     % t_n &= t^A_g\\[\vd]
%     % t_{i+1} &= t_i + w_i + \tau(e_{v_i v_{i+1}}, m_i) \quad\forall~i\in[1, n-1]\\[\vd]
%     % % t^D_i &= t^A_i + w_i \quad\forall~i\in\mathcal{P}\\[\vd]
%     % w_i &\leq \overline{w}_{v_i} \qquad\quad\forall~i\in[1,n-1]\\[\vd]
%     % t^A_{i+1} &= t^D_i + \tau(\hat{e}_i,1)~\mu(\hat{e}_i,t^D_i)\\ &\qquad\qquad + \tau(\hat{e}_i,0)~(1-\mu(\hat{e}_i,t^D_i))\\
%     % &\qquad\qquad\qquad\forall~i\in\mathcal{P}.
% \end{align*}
% Here $\mathcal{P}.v_1$ and $\mathcal{P}.t_1$ is the vertex and arrival time of the first element in path $\mathcal{P}$.

To efficiently solve this problem, we must make three crucial decisions: selecting edges to travel, choosing the mode of operation, and determining the waiting time at each vertex. Our proposed method offers a novel approach to computing the optimal solution. However, before delving into the details of our solution, it is necessary to grasp the concept of budget and how new nodes are generated during the search process.
% % 
% Here, $\hat{e}_i$ is the edge connecting vertices $v_i$ and $v_{i+1}$ in path $\mathcal{P}$, and $\mu(\hat{e}_i,t^D_i)$ is the mode of operation used on edge $(v_i,v_{i+1})$. 

{
}

\section{Budget and Node Generation} \label{sec:budget}
% A common way of solving time-dependent graph search problems is to augment the vertices with an added dimension of time and perform the search in this higher dimensional space.  
Since the robot is allowed to wait (subject to the waiting limits), it is possible to delay the robot's arrival at a vertex by waiting at one or more of the preceding vertices.  Moreover, the maximum amount of time by which the arrival can be delayed at a particular vertex depends on the path taken from the start to that vertex.  Our key insight is that this information about the maximum delay can be used to efficiently solve the given problem by removing the need to examine the vertices at every possible arrival time.
We achieve this by augmenting the search space into a higher dimension, using additional parameters with the vertices of the given graph.
A node in our search is defined as a triplet $(x,a_x,b_x)$, corresponding to a vertex $x \in V$, arrival time $a_x \in \mathbb{Z}_{\geq 0}$ and a budget $b_x \in \mathbb{Z}_{\geq 0}$.  The \textit{budget} here defines the maximum amount of time by which the arrival at the given vertex can be delayed.  Thus, the notion of budget allows a single node $(x,a_x,b_x)$ to represent a range of arrival times from $[a_x, a_x+b_x]$ at vertex $x$.  Therefore, the allowed departure time from this vertex lies in the interval $[a_x, a_x+b_x+\overline{w}_x]$.

% \begin{figure}
%     \centering
%     \includegraphics[width=0.9\columnwidth]{Figures/Budget.png}
%     \caption{An example explaining the notion of budget and determining the resulting node when a node $(x, a_x, b_x)$ is explored. $t_{A1}$ and $t_{A2}$ are, respectively, the minimum and maximum arrival times at vertex $y$ under each mode of operation. The arrival time at the resulting node corresponds to the earliest departure time from $x$, while the resulting budget gets increased or decreased depending on the amount of waiting done at $x$ and the time of change in operator availability.}
%     \label{fig:budget}
% \end{figure}

% \begin{figure}
%     \centering
%     \includegraphics[width=0.9\columnwidth]{Figures/budget_graph.png}
%     \caption{Relationship between departure time $d_x$ from vertex $x$ and arrival time $a_y$ at vertex $y$.  These parameters come from the previous figure.}
%     \label{fig:budgetGraph}
% \end{figure}

\subsection{Node Generation}
The proposed algorithm is similar to standard graph search algorithms, where we maintain a priority search queue, with nodes prioritized based on the earliest arrival time (plus any admissible heuristic).  Nodes are then extracted from the queue, their \textit{neighbouring} nodes are generated and are added to the queue based on their priority.
Since in our search a node is defined by the vertex, arrival time and budget, we must determine these parameters for the newly generated nodes when exploring a given node.
To characterize the set of nodes to be generated during the graph search in our proposed algorithm, we define the notion of direct reachability as follows.

\begin{definition}[Direct reachability]
A node $(y, a_y, b_y)$ is said to be \emph{directly reachable} from a node $(x,a_x,b_x)$ if $x$ and $y$ are connected by an edge, i.e., $e_{xy}\in E$, and it is possible to achieve all arrivals times in $[a_y, a_y+b_y]$ at $y$ through edge $e_{xy}$ for some departure time $t_D \in [a_x, a_x+b_x+\overline{w}_x]$ from $x$ and some mode of travel.
\end{definition}

As an example, consider a node $(x,10,2)$ with $\tau(e_{xy},0) = 5$ and $\overline{w}_x=3$.  Then the nodes $(y, 15, 5), (y, 16, 4)$ and $(y, 17, 3)$ are a few directly reachable nodes from $(x,10,2)$ (corresponding to departure times $10$, $11$ and $12$, respectively).  %In other words, if the arrival time at $x$ is in the range $[10, 12]$, then it is possible to arrive at $y$ at any time in $[15, 20]$.  The nodes $(y, 16, 4), (y, 17, 3)$, are some other directly reachable nodes from $(x,10,2)$ (corresponding to departure times $11$ and $12$, respectively).
\begin{figure}
    \centering
    \includegraphics[width=0.95\columnwidth]{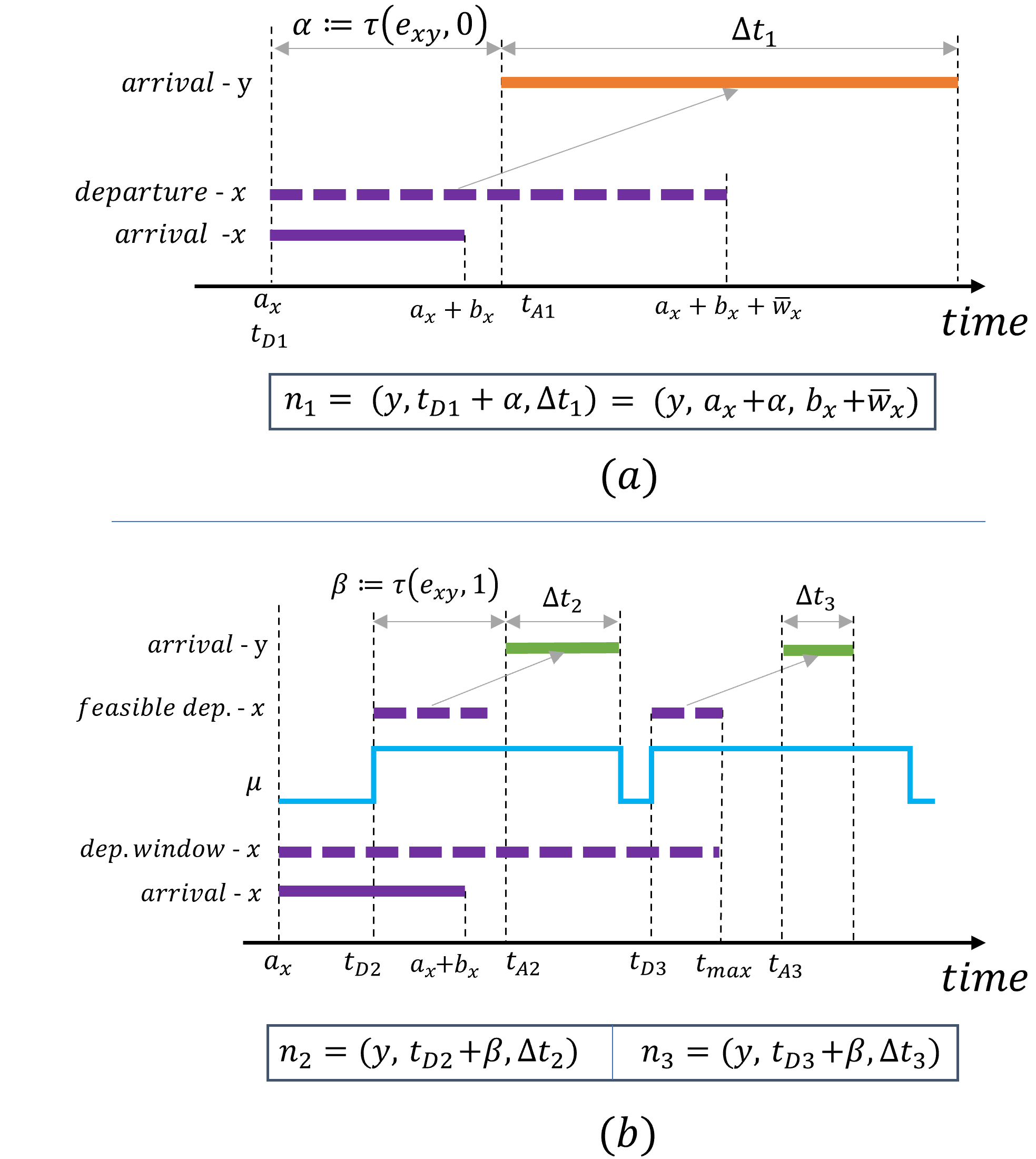}
    \caption{An example explaining the notion of budget and determining the resulting node when a node $(x, a_x, b_x)$ is explored.  (a) Under autonomous mode, the earliest departure from $x$ is $a_x$ (i.e., no waiting).  Since autonomous mode is always available to the robot, the corresponding arrival time at $y$ forms a single block, shown as a solid orange line.  This can be thus represented using a single node $n_1$.  (b) Under the assisted mode, feasible departure times from $x$ are governed by the operator availability function $\mu$, and a time $t_{max}=\min(\alpha-\beta, b_x, \overline{w}_x)$.  In this example, the resulting arrival time at $y$ forms two separate blocks, shown as solid green lines on the time axis.  Thus, exploring the node under assisted operation generates two nodes, $n_2$ and $n_3$.\vspace{-2.0ex}} 
    \label{fig:newBudget}
\end{figure}

Like standard graph search methods, our algorithm aims to generate all nodes directly reachable from the current node during the exploration process.  One approach is to generate all directly reachable nodes from the given node $(x,a_x,b_x)$ for all possible departure times in $[a_x, a_x+b_x+\overline{w}_x]$.  However, this results in redundancy when multiple nodes can be represented collectively using a single node with a suitable budget. 

% For a given node $(x,a,b)$ and autonomous mode of operation, there is only one critical departure time, which is the same as $a$ the arrival time at $x$.  This is because there are no times when the use of autonomous mode is restricted. 
% For the assisted mode, critical departure times are when the operator becomes available to assist for the given edge.
% % A critical time for an edge $e_{uv}$ is defined as a departure time from vertex $u$ such that the arrival time at vertex $v$ is discontinuous with respect to the departure time.
% In the example shown in Fig.~\ref{fig:budget}, there are two critical times $t$ and $t+\Delta t_1$.  Exploring the node at these two departure times gives us two nodes for vertex $y$ that cover the entire range of arrival times achievable from the node $(x,a_x,b_x)$.

% \begin{proposition}
%     Critical times are departure times from a given vertex that are sufficient to generate enclosing nodes for all vertex-time pairs that are directly reachable from the current node to the given neighbor.  
% \end{proposition}

As the operator availability changes, the possible arrival times at the next vertex may present themselves as separate blocks of time.  A block of arrival times can be represented using a single node, and thus we only need to generate a new node for each arrival time block.  To understand this, we consider the example given in Fig.~\ref{fig:newBudget}, where a node $(x,a_x,b_x)$ is being extracted from the queue, and we want to generate the nodes corresponding to a neighbouring vertex $y$.  The arrival time range at $x$, $[a_x, a_x+b_x]$ is shown as solid purple line.  The possible departure window $[a_x, a_x+b_x+\overline{w}_x]$ is shown as purple dashed line. 

Under autonomous operation, the edge can be traversed by departing at any time in the departure window, resulting in possible arrival time at vertex $y$ in the interval $[a_x+\alpha, a_x+b_x+\overline{w}_x+\alpha]$, shown as solid orange line.  Therefore, we can represent these possible arrival times using the node $n_1 = (y, a_x + \alpha, b_x+\overline{w}_x)$, where $ a_x + \alpha$ is the earliest arrival time at $y$ and $b_x+\overline{w}_x$ is the new budget.  Note that the new budget is increased from the previous value by an amount of $\overline{w}_x$.

Under assisted operation, only a subset of departure window is feasible, as shown in Fig.~\ref{fig:newBudget}(b).  This results in separate blocks of arrival times at $y$, shown as solid green lines.  The range of arrival times corresponding to these blocks become the budget values for the new nodes.  In the given example, two nodes are generated: $n_2 = (y, t_{D2}+\beta, \Delta t_2)$ and $n_3 = (y, t_{D3}+\beta, \Delta t_3)$.

% We can identify the earliest time that the edge $e_{xy}$ can be traversed with operator assistance.  In the example given in Fig.~\ref{fig:newBudget}(b), this happens at time time $t_{D2}$.  For assisted mode, the operator must be available throughout the entire edge duration $\beta$. We represent this condition as $\mu([t_{D2},t_{D2}+\beta]) = 1$, and if it is met, we can use assisted mode to traverse the edge.
% We define $\Delta t_2$ as the time difference between the edge's earliest completion time $t_{A2}$ and the time when operator availability $\mu$ changes to $0$.  The assisted operation is valid for all departure times in the range $[t_{D2} + \Delta t_2]$.  Therefore, we can represent the resulting block of possible arrival times using the node $n_2 = (y, t_{D2}+\beta, \Delta t_2)$.  
% Now we consider the time $t_{D3}$, the next time when the edge can be travelled under assisted mode.  However, in this case (for this particular example), we can only delay the departure by an amount $\Delta t_3$, as $t_{D3} + \Delta t_3 = \mathcal{T}_D$, the latest departure time from $x$ that we need to consider.  Thus we generate a new node $n_3 = (y, t_{D3}+\beta, \Delta t_3)$.

Note that the feasible departure times are limited by operator availability and $t_{max}$.  The value of $t_{max}$ is the minimum of $b_x + \overline{w}_x$ and $\alpha - \beta$. The former quantity limits the departure from $x$ to $a_x+b_x+\overline{w}_x$, while the latter 
comes from the observation that any departure time $t_{D} > a_x + (\alpha - \beta)$ will result in arrival at $y$ at a time $a_y > a_x + \alpha$, and a budget $b_y < b_x + \overline{w}_x$.  However, this arrival time range is already covered by the node generated under autonomous operation.

% Note that the first block of feasible departure times is due to the operator availability becoming $0$, while the second block is limited by $t_{max}$.  The quantity $t_{max}$ is computed as the minimum of two quantities: 1) $b_x + \overline{w}_x$, the maximum delay in departure from $x$, and 2) $\alpha-\beta$, the difference between task durations under the two modes.  The first quantity reflects the requirement that departure from $x$ cannot exceed $a_x+b_x+\overline{w}_x$.  The second quantity comes from the observation that any departure time $t_{D} > a_x + (\alpha - \beta)$ will result in arrival at $y$ at a time $a_y > a_x + \alpha$, and a budget $b_y < b_x + \overline{w}_x$.  Recall that, under autonomous mode, the node $(y, a_x + \alpha, b_x+\overline{w}_x)$ is generated.  This node has an earlier arrival time and a larger budget, and therefore already includes the arrival time block that would have resulted by any such departure time $t_{D}$.  Thus $t_{max}$ takes into account both the remaining budget from the previous node and the maximum useful delay in departure.

% Note that the entire range of possible arrival times at $y$ is captured using just three nodes $n_1, n_2$ and $n_3$.  
\emph{Critical departure times:} Note that the earliest arrival times for each of the three new nodes correspond to unique departure times from $x$ ($t_{D1}, t_{D2}, t_{D3}$ in Fig.~\ref{fig:newBudget}).  We refer to these times as {critical departure times}, as exploring a node only at these times is sufficient to generate nodes that cover all possible arrival times at the next vertex.
Since in the presented problem, the edge duration depends on the mode of operation selected, the set of critical departure times for a node is a subset of times when the operator availability changes, and thus can be efficiently determined.  
% For all possible departures between two critical times and a given mode of operation, the arrival times at the next vertex are present as a single block of time and thus can be represented using a single node.

Next, we present how these concepts are used by our proposed Budget-$A^*$ algorithm to solve the given problem.

\section{Budget $A^*$ Algorithm}
This section details the proposed Budget $A^*$ algorithm and its three constituent functions: EXPLORE, REFINE and GET-PATH.
To recall, a node in our search is defined as a tuple $(x,a_x,b_x)$. %, denoting the vertex $x \in V$, arrival time $a_x \in \mathbb{Z}_{\geq 0}$ and a budget $b_x \in \mathbb{Z}_{\geq 0}$.  
A pseudo-code for the Budget-$A^*$ algorithm is given in Alg.~\ref{alg:budget_dij}, and more details on the constituent functions follow.  
\begin{algorithm}
\begin{algorithmic}[1]
\STATE \textbf{Input:} $G = (V, E), \tau, \mu, \overline{w}, s, g$
\STATE $Q \gets \text{initialize priority queue}$
\STATE $S \gets \emptyset$
\STATE $Q.\text{push}((s, 0, 0))$
\STATE $\psi(x,a) \gets nil$ for all $x\in V, a \in \mathbb{Z}_{\geq0}$ // Initialize predecessor function
\WHILE{$Q$ not empty}
\STATE $\nar{currNode} \coloneqq (x, a_x, b_x) \gets Q.\text{extract-min}()$ \label{line:extractMin}
\STATE $S \gets S \cup \nar{currNode}$
\IF{$x = g$}
\STATE \textbf{break}
\ENDIF
\FORALL{$y \in \text{neighbors}(x)$} \label{line:eachNeigh}
\STATE $\mathcal{N} \gets \text{EXPLORE}(a_x, b_x, \overline{w}_x, e_{xy}, \tau, \mu)$
% \STATE $\mathcal{N} \gets \text{EXPLORE}(a_x, b_x, \overline{w}_x, \alpha, \beta, \mu)$
\FORALL{$(y, a_i, b_i, m_i) \in \mathcal{N}$}
\STATE $\nar{newNode} \gets (y, a_i, b_i)$
\STATE $Q, \psi \gets \text{REFINE}(Q, \nar{newNode}, \nar{currNode}, \psi, m_i)$
\ENDFOR
\ENDFOR  \label{line:endForEachNeigh}
\ENDWHILE
\STATE $path \gets \text{GET-PATH}(\nar{currNode}, \overline{w}, \psi, \tau)$
\end{algorithmic}
\caption{Budget $A^*$}
\label{alg:budget_dij}
\end{algorithm}
The algorithm initializes an empty priority queue $Q$, a processed set $S$ and a predecessor function $\psi$.  
It then adds a node $(s,0,0)$ to $Q$ denoting an arrival time of exactly $0$ at $s$.  The algorithm iteratively extracts the node with the earliest arrival time (plus an admissible heuristic) from $Q$, adds it to $S$, and generates new candidate nodes for each of its neighbors using the EXPLORE function.  The REFINE function then checks if these nodes can be added to the queue, removes redundant nodes from $Q$, and updates predecessor information.  The algorithm continues until $Q$ is empty or the goal vertex is reached. The GET-PATH function generates the required path using the predecessor data and waiting limits. % Alg.~\ref{alg:budget_dij} provides the pseudo-code for the algorithm.
% 

%%%%%%%%%%%%%%%%%%%%%%%%%%%%%%%%%%%%%%%%%%%%%%%%%%%%%%
% Budget
%%%%%%%%%%%%%%%%%%%%%%%%%%%%%%%%%%%%%%%%%%%%%%%
% EXPLORE
\subsection{Exploration}
% In this step, we generate a list of potential nodes to add to the search queue.
% The EXPLORE function takes in several input parameters: arrival time $a_x$, budget $b_x$, waiting limit $\overline{w}_x$, autonomous and assisted travel durations $\alpha$ and $\beta$, and operator availability $\mu$.  
The EXPLORE function takes in several input parameters: arrival time $a_x$, budget $b_x$, waiting limit $\overline{w}_x$, edge $e_{xy}$, travel durations $\tau$ and operator availability $\mu$.  The function returns a set $\mathcal{N}$ of candidate nodes of the form $(y, a_i, b_i, m_i)$, where $a_i,b_i,m_i$ are the arrival time, budget and the mode of operation respectively, corresponding to all critical departure times from the node $(x, a_x, b_x)$ to vertex $y$.  A pseudo-code is shown in Alg.~\ref{alg:explore}.

As discussed in Sec.~\ref{sec:budget}, the autonomous mode generates one new node, while assisted mode can generate multiple nodes depending on operator availability, node budget and task duration.  The function first adds a node $(y,a_x+\alpha, b_x+\overline{w}_x)$ corresponding to the autonomous mode to $\mathcal{N}$.  % The node is $(y,a_x+\alpha, b_x+\overline{w}_x)$, where $\alpha = \tau(e_{xy}, 0)$. % is the autonomous duration of edge $e_{xy}$.  

For the assisted mode, it first computes the maximum useful delay in departure $t_{max}$.  Next, it generates an ordered set $\mathcal{F}$ of feasible departure times from the current node as the times in departure window when it's possible to depart under assisted mode, computed using $\mu$ and $\beta$ (line~\ref{line:feasible}). 
% To find the critical departure times, which are the times when operator becomes available to assist for the edge, we use 
Lines~\ref{line:td-1}-\ref{line:addNewNode} generate a new node $(y, a_y, b_y)$ for each critical departure time $t_d$, with a budget $b_y = 0$ and arrival time $a_y = t_d + \beta$.  The budget is then incremented for each consecutive departure time in $\mathcal{F}$.  A gap in $\mathcal{F}$ means a gap in arrival time at $y$ indicating that we have considered the complete arrival time range for that critical departure time.  This condition is checked in line~\ref{line:td+1}, and the node $(y,a_y, b_y, 1)$ is added to $\mathcal{N}$.
\begin{algorithm}
\begin{algorithmic}[1]
% \STATE \textbf{Input:} Arrival time $t$, budget $b$, autonomous travel time $t_a$, assisted travel time $t_t$, and operator availability $\mu$ for an edge $(u,v)$
\STATE $\alpha \gets \tau(e_{xy}, 0)$, $\beta \gets \tau(e_{xy}, 1)$
\STATE $\mathcal{N} \gets \{(y, a_x+\alpha, b_x + \overline{w}_x, 0)\}$ %// to account for autonomous edge
\STATE $t_{max} \gets \min(\alpha - \beta, b_x + \overline{w}_x)$ \label{line:tMax}
\STATE $\mathcal{T}_D \gets [a_x, a_x + t_{max}]$ // Possible departure window
\STATE $\mathcal{F} \coloneqq [t \in \mathcal{T}_D\;~s.t.~\; \mu([t, t+\beta]) = 1]$ // Feasible set \label{line:feasible}
\FORALL {$t_d \in \mathcal{F}$} \label{line:forallTd}

\IF {$t_d - 1 \notin \mathcal{F}$} \label{line:td-1}
\STATE $(y, a_y, b_y) \gets (y, t_d + \beta, 0)$ \label{line:addNewNode} %// Critical departure time $t_d$
\ELSE
\STATE $b_y \gets b_y + 1$
\ENDIF
\IF {$t_d + 1 \notin \mathcal{F}$} \label{line:td+1}
\STATE $\mathcal{N} \gets \mathcal{N} \cup (y, a_y, b_y, 1)$
\ENDIF
\ENDFOR \label{line:allTdEnd}
\RETURN $\mathcal{N}$
\end{algorithmic}
% \caption{EXPLORE$(a_x, b_x, \overline{w}_x, \alpha, \beta, \mu)$}
\caption{EXPLORE$(a_x, b_x, \overline{w}_x, e_{xy}, \tau, \mu)$
}
\label{alg:explore}
\end{algorithm}
% \vspace{-1.5ex}

Once all departure times in $\mathcal{F}$ are accounted for, the set $\mathcal{N}$ contains all required arrival time and budget pairs (along with the mode of operation) for the given node $(x,a_x,b_x)$ and neighbour $y$. % while accounting for the operator availability, the travel time and the waiting limit.

\subsection{Node Refinement}
% After the set of directly reachable nodes is generated, the REFINE function determines if the newly generated nodes should be added to the priority queue $Q$ and if any of the existing nodes can be removed. The function verifies if the time range in which the new node is able to arrive at its vertex is entirely included within the arrival time window of any existing node in the queue $Q$ that has the same vertex.  This is accomplished using the check on line~\ref{line:refine1} of Algorithm~\ref{alg:refine}. If the condition is met for a node, it can be disregarded as it offers no new information. However, if the condition is not met, the new node is added to $Q$ and a further check is performed to identify any nodes in $Q$ whose arrival time window is fully contained within the new node's window (line~\ref{line:refine2}). These redundant nodes are removed from $Q$.
The REFINE function determines which nodes to add or remove from the search queue, based on the newly generated nodes. 
The function checks if the new node is redundant by comparing its vertex and arrival time window with nodes already in the queue (Alg.~\ref{alg:refine} line~\ref{line:refine1}).  If the new node is found to be redundant, the function returns the original queue without modifications.  If not, the new node is added to the queue, and if there is any node in $Q$ with the same vertex and an arrival time range subset of the new node's range (line~\ref{line:refine2}), it is removed.  The function then returns the updated queue and predecessor function.
 
\begin{algorithm}
\begin{algorithmic}[1]
\STATE $\nar{toRemove} \gets \emptyset$
\STATE $(y,a_y,b_y) \gets \nar{newNode}$
\STATE $(u,a_u,b_u) \gets \nar{currNode}$
\FORALL{$(x,a,b) \in Q$}
\IF{$x = y$ and $a \leq a_y$ and $a+b \geq a_y+b_y$} \label{line:refine1}
\RETURN $Q, \psi$
\ELSIF{$x = y$ and $a \geq a_y$ and $a+b \leq a_y+b_y$} \label{line:refine2}
\STATE $\nar{toRemove} \gets \nar{toRemove} \cup {(x,a,b)}$
\ENDIF
\ENDFOR
\STATE $Q \gets Q \setminus \nar{toRemove}$
\STATE $Q \gets Q \cup \nar{newNode}$
\STATE $\psi(y,a_y) = (u, a_u, m)$
\RETURN $Q, \psi$
\end{algorithmic}
\caption{REFINE$(Q, \nar{newNode}, \nar{currNode}, \psi, m)$}
\label{alg:refine}
\end{algorithm}
\vspace{-2.0ex}

%%%%%%%%%%%%%%%%%%%%%%%%%%%%%%%%%%%%%%%%%%%%%%
% GET PATH
\subsection{Path Generation} \label{sec:getPath}
To get the execution path from start to goal, we use the predecessor data stored in function $\psi$, which returns the predecessor node vertex and arrival time $(x,a_x)$ along with the mode of travel $m_x$, used on the edge $e_{xy}$ for a given vertex-time pair $(y,a)$.
% 
% When we extract the goal vertex from the priority queue, we can use predecessor data to get the execution path from start to goal.  For a given vertex-time pair $(y,a)$, the function $\psi(y,a)$ returns the predecessor node vertex and arrival time $(x,a_x)$ along with $m_x$, the mode of travel used for traversing the edge $e_{xy}$.   % , i.e., the sequence of edges, the waiting times at each intermediate vertex and the mode of operation chosen for each edge. 
% Since, the problem does not follow the optimal substructure property, we cannot just save only the predecessors vertices as the nodes are created.  We need to know the exact path from which the goal vertex is reached.  This can be done by storing both predecessor vertex and its arrival time (because a vertex and arrival time pair is unique in our search).
However, we need to determine the exact arrival and departure times at each vertex based on wait limits. To achieve this, we use the GET-PATH function shown in Alg.~\ref{alg:get_path}.  The function backtracks from the goal to the start, calculating the exact departure time from the predecessor vertex based on the earliest arrival time at the current vertex and the mode of operation (line~\ref{line:tD_path}). 
The exact arrival time is then determined using the departure times and the maximum allowed waiting $\overline{w}$ (lines~\ref{line:tD_path}-\ref{line:tA_path}).   The final path is stored as a list of tuples representing a vertex, the arrival time, waiting time, and mode of operation used.

\begin{algorithm}
\begin{algorithmic}[1]
% \STATE \textbf{Input:} Processed set $S$, max wait times $\overline{w}$
\STATE Initialize an empty path list $\mathcal{P}$
% \STATE $(y,a,b) \gets$ the last element of $S$
\STATE Append $(y, a, 0, 0)$ to $\mathcal{P}$
\STATE $\nar{rem} \gets 0$
\WHILE{$\psi(y,a) \neq nil$}
\STATE $(x, a', m) \gets \psi(y,a)$
\STATE $t_D \gets a - m~\tau(e_{xy}, 1) - (1-m)\tau(e_{xy}, 0) + \nar{rem}$ \label{line:tD_path}
% \STATE $p \gets node | node \in S, node.u == c.p$
\STATE $w_x \gets min(t_D - a', \overline{w}_{x})$ \label{line:w_path}
\STATE $\nar{rem} \gets t_D - a' - w_x$
\STATE $a_x \gets a' + \nar{rem}$ // Required arrival at $x$ \label{line:tA_path}
\STATE Append $(x, a_x, w_x, m)$ to $\mathcal{P}$
\STATE $(y, a) \gets (x, a')$
% \STATE $(x,a',m) \gets \psi(x,a')$
% \STATE $c \gets p$
\ENDWHILE
\RETURN $\mathcal{P}$ 
\end{algorithmic}
\caption{$GET\_PATH((y, a, b),\overline{w}, \psi, \tau)$}
\label{alg:get_path}
\end{algorithm}
\vspace{-2.0ex}
%%%%%%%%%%%%%%%%%%%%%%%%%%%%%%%%%%%%%%%%%%%%%%%%%%%%%%%%%%%%%%%%%%%%%%%%%%%%%%%%%%%%%%%%
{
}
%%%%%%%%%%%%%%%%%%%%%%%%%%%%%%%%%%%%%%%%%%%%%%

\subsection{Correctness Proof}
Let a vertex-time pair $(y, a_y)$ be called directly reachable from a node $(x,a_x,b_x)$, if vertex $y$ can be reached by departing vertex $x$ between $a_x$ and $a_x+b_x+\overline{w}_x$ through edge $e_{xy} \in E$.
\begin{lemma}
Consider a node $(x,a_x,b_x)$ extracted from $Q$ (line \ref{line:extractMin}), and a $y \in \text{neighbors}(x)$ inspected in lines \ref{line:eachNeigh}-\ref{line:endForEachNeigh}.  If $(y, a_y)$ is directly reachable from $(x,a_x,b_x)$, then there is a node in $Q$ with vertex $y$ whose arrival time range contains $a_y$.  
% exists at least one node in $Q$ for every achievable arrival time $a_y$ at $y$ (when departing $x$ between $a_x$ and $a_x+b_x+\overline{w}_x$), such that the arrival time range of the node includes $a_y$. %, there exists at least one node $(v,a',b')$ present in $Q$.
\label{th:directReach}
\end{lemma}

\begin{proof}
% The proof follows from the description of budget computation in Sec.~\ref{sec:budget} and definition of critical departure times.
% What we want to show here is that EXPLORE function generates nodes for every block of arrival times at the next vertex, and then the REFINE function only removes nodes for which the arrival time window/range/block is already covered by another node.

For a given node, the critical departure times represent the number of separate arrival time blocks.  Also, as discussed earlier, a single block of arrival times can be represented by a node having the earliest arrival time in that block and budget equal to the width of the block.  The EXPLORE function gets called for each neighbour of $x$ (Alg.~\ref{alg:budget_dij} line~\ref{line:eachNeigh}) and generates new nodes corresponding to each critical departure (Alg.~\ref{alg:explore} lines~\ref{line:forallTd}-\ref{line:allTdEnd}).  Therefore the resulting nodes cover all possible arrival times at every neighbouring vertex of $x$ when departing at a time in the range $[a_x, a_x + b_x + \overline{w}_x]$. 

During the refinement step, only those nodes are removed for which the arrival time range is already covered by another node (Alg.~\ref{alg:refine} line~\ref{line:refine2}). 
Therefore, after execution of the EXPLORE and REFINE functions, there exist nodes for all achievable arrival times at the neighboring vertices corresponding to the node $(x,a_x,b_x)$.
\end{proof} 
\begin{lemma}
    When a node $(x,a_x,b_x)$ is extracted from $Q$, for every achievable arrival time $a' < a_x$ at $x$ (through any path from the start vertex), there exists at least one node with vertex $x$ in the explored set $S$ for which the arrival time range includes $a'$.
    \label{th:noEarlierArrival}
\end{lemma}
\begin{proof}
We will use proof by induction. 
Base case: Consider the starting node $(s,0,0)$ (first node extracted from $Q$). Since it has an arrival time of $0$, and arrival times are non-negative there is no earlier achievable arrival time at vertex $s$, so the statement is true.

Induction step: Assume the statement is true for the first $k$ nodes extracted and added to $S$.  We want to show that it is also true for the next node $(x,a_x,b_x)$ extracted from $Q$. We will prove this by contradiction.

Suppose there exists an achievable arrival time $a'<a_x$ at $x$ such that no node of vertex $x$ in $S$ has an arrival time range that includes $a'$.  Let $(x,a')$ is achieved via some path\footnote{Here, only the vertex-time pairs are used to denote a path. Wait times and mode of travel are omitted for simplicity.} $(s,0) \leadsto (u,a_u) \rightarrow (v,a_v) \leadsto (x,a')$, where $(u,a_u)$ and $(v,a_v)$ are two consecutive entries in the path.  Let $(v,a_v)$ be the first pair in the path for which a node enclosing arrival time $a_v$ is not present in $S$.  This can also be $(x,a')$ itself.  
Since $(v,a_v)$ is directly reachable from $(u, a_u)$, when exploring the node corresponding to $(u,a_u)$, a node corresponding to arrival time $a_v$ at $v$ must have been inserted (or already present) in $Q$ (Lemma~\ref{th:directReach}).  Let this node be $(v,a'_v, b'_v)$.

We have $a_v \in [a'_v, a'_v + b'_v]$.  Since $b'_v \geq 0$, we get $a'_v \leq a_v$.  Also, $a_v \leq a'$ because $(v,a_v)$ and $(x,a')$ lie on a valid path.  Since we assumed $a' < a_x$, we get $a'_v < a_x$.  However, since $(x, a_x, b_x)$ is extracted from $Q$ first, we must have $a_x \leq a'_v$.  Therefore, the initial assumption must be incorrect, and the statement holds for any node extracted from $Q$.
\end{proof}

%%%%%%%%%%%%%%%%%
%%%%%%%%%%%%%%%%%

\begin{theorem}
Consider a vertex $x$, and let $(x,a_x,b_x)$ be the first node with vertex $x$ that is extracted from $Q$.  Then $a$ is the earliest achievable arrival time at $x$.
\end{theorem}

\begin{proof}
By Lemma~\ref{th:noEarlierArrival}, if there exists an arrival time $a' < a_x$ at $x$ which is achievable through any path from the start, a corresponding node must be in the explored set. Since $(x,a_x,b_x)$ is the first node with vertex $x$ that is extracted from $Q$, there is no node with vertex $x$ in the explored set $S$. Therefore, $a$ is the earliest achievable arrival time at $x$.
\end{proof}

\section{Simulations and Results} \label{sec:sim}
In this section, we present the simulation setup and discuss the performance of different solution methods.  % To demonstrate the performance of our algorithm, we take the example of last-mile robot delivery in a city, where streets can be characterized with different speeds and waiting rules.
\subsection{Baseline Algorithms}
% The presented problem can be solved using some of the existing methods for time-dependent shortest path problem with some modifications.  
In this section, we present some solution approaches that we use to compare against the proposed Budget-$A^*$ algorithm.

\subsubsection{TCSP-CWT} 
The TCSP-CWT algorithm (Time-varying Constrained Shortest Path with Constrained Waiting Times), presented in \cite{cai1997time}, solves the shortest path problem under the constraint of a bounded total travel time.  To solve the given problem, we modify the original graph by creating two copies of each vertex, one for autonomous mode and another for assisted mode.  New edges are added accordingly.  The search is stopped at the first time step with a finite arrival time at the goal vertex.

\subsubsection{Time-expanded $A^*$} 
The Time-expanded $A^*$ algorithm is a modified version of the $A^*$ algorithm that can be used to solve the given problem  \cite{ford1958constructing}. It creates a separate node for each vertex at each time step, and adds new edges based on the waiting limits and operator availability. %However, this method scales poorly with increasing number of vertices and the time horizon.

\subsubsection{Greedy (Fastest Mode) Method}
One efficient method for obtaining a solution is to combine a time-dependent greedy selection with a static graph search method. 
This approach is similar to an $A^*$ search on a static graph, but takes into account the arrival time at each vertex while exploring it. To determine the edge duration to the neighboring vertices, we consider the faster of the two alternatives: traversing the edge immediately under autonomous mode or waiting for the operator to become available. Once the goal vertex is extracted from the priority queue, we can stop the search and use the predecessor data to obtain the path. 
\subsection{Problem instance generation}
\begin{figure}
    \centering
    \includegraphics[width=0.6\columnwidth]{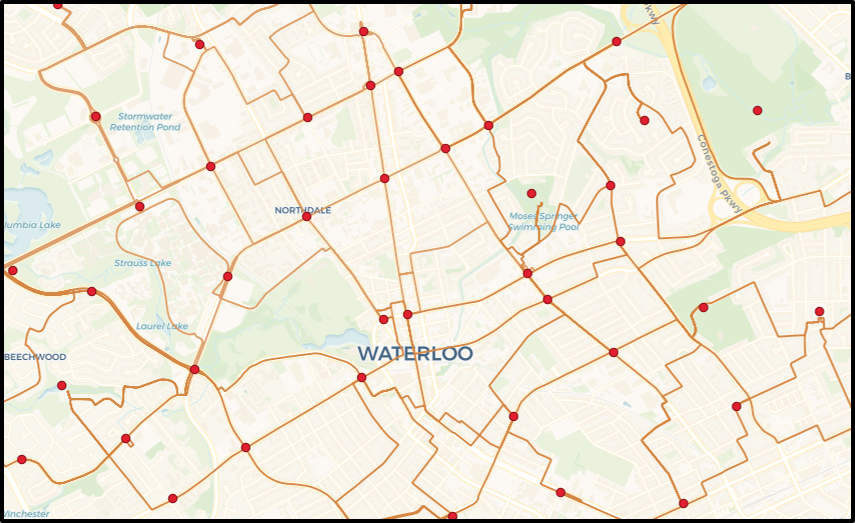}
    \caption{Example of the street network graph of the city of Waterloo used in the simulations.  The figure shows a screenshot from the QGIS software.  The shortest paths (orange lines) between vertices (red dots) are generated using the OpenRouteService.\vspace{-2.5ex}  %Higher number of vertices allow for a finer specification of robot speeds and waiting limits at different location in the map.
    }
    \label{fig:graphMap}
\end{figure}

% In our simulations, we obtain a distance graph corresponding to an actual city road network using real-world data from a Geographic Information System (GIS).  After the distance data is obtained, we then need to specify the robot's speed under autonomous and assisted operation, as well as the allowed maximum waiting duration.  In practice, these specifications are provided by system designers based on robot-related and environment-related factors, and are considered as an input for the planning problem. %, we chose to create large random graphs resembling a city street network, and randomly assign waiting limits and travel durations along the edges. An example of such a map is shown in Figure~\ref{fig:graphMap}.
% 
For generating the problem instances, we use the map of the city of Waterloo, Ontario, Canada (a $10km \times 10km$ area around the city centre).  Using the open source tools QGIS and OpenStreetMap, we place a given number of points at different intersections and landmarks.  These points serve as vertices in our graph.  Next, we use Delaunay triangulation to connect these vertices and use OpenRouteService (ORS) to compute the shortest driving distance between these vertices.
% An edge is added between two vertices if the shortest path from one vertex to the other does not pass through any other vertex in the graph. 
% using the Delaunay Triangulation method. 
An example graph of the city is shown in Figure~\ref{fig:graphMap}.
% We then assign waiting limits, travel durations and determine operator availability, representing the time-dependent nature of the problem. 
To obtain the travel durations at each edge, we first sample robot speeds from a uniform random distribution.  The travel durations under the two modes %: $\alpha_{xy} \coloneqq \tau(e_{xy}, 0)$ and $\beta_{xy} \coloneqq \tau(e_{xy}, 1)$ 
are then computed by dividing the edge length (computed using ORS) by the speed values and rounding off to the nearest integer.  The travel speeds are sampled as follows: autonomous speed $u^0_{xy} \sim U[0,40]$; assisted speed $u^1_{xy} \sim  U[10 + u^0_{xy},30 + u^0_{xy}]$.
%$u^1_{xy} \gets u^0_{xy} + \Delta u_{xy}$. %One is used as the travel speed under autonomous mode ($u^0_{xy}$), and the sum of two is used as the travel speed under assisted mode ($u^1_{xy}$), for each edge $e_{xy}\in E$.  .
% \begin{align*}
% \centering
% u^0_{xy} &\sim U[10,50], \;\; \Delta u_{xy} \sim U[10,30],
% u^1_{xy} \gets u^0_{xy} + \Delta u_{xy}.
% % \alpha_{xy} &\gets d_{xy}/u^1_{xy}, \;\;
% % \beta_{xy} \gets d_{xy}/u^0_{xy}.
% \label{eq:sample_time}
% \end{align*}
The maximum waiting duration at each vertex $x$ is sampled from a uniform random distribution as $\overline{w}_x \sim U[0,15]$.  
% \begin{align*}
% \overline{w}_x \sim U[0,5].
% \end{align*}
The operator availability function is generated by randomly sampling periods of availability and unavailability, with durations of each period sampled from the range of $[10, 200]$. 
The distance values used in our simulations are in meters, times are in minutes and speeds are in meters/minute.
We test the algorithms using the Waterloo city map with varying vertex density, by selecting $64$, $100$ or $225$ vertices to be placed in the map.  We generate $20$ problem instances for each density level (varying speeds, waiting limits and operator availability), and for each instance, we solve the problem for $100$ randomly selected pairs of start and goal vertices.  The algorithms are compared based on solution time and the number of explored nodes.  We also examine some of the solutions provided by the greedy method.

\textbf{Note on implementation:} 
All three graph search algorithms (Budget-$A^*$, Greedy and Time-expanded $A^*$) use the same heuristic, obtained by solving a problem instance under the assumption that operator is always available.  This heuristic is admissible in a time-dependent graph \cite{zhao2008algorithm} and can be computed efficiently.  The priority queues used in all methods are implemented as binary heaps, allowing for efficient insertion, extraction and search operations.  Additionally, all the methods require computation of the feasibility set (Alg.~\ref{alg:explore} line~\ref{line:feasible}).  This is pre-computed for all departure times and is given as input to each algorithm. 

% \subsection{Application and performance}
% Here we show how this method can solve the problem and give us the path to take, the mode to use and the durations to wait at each intermediate step.
\subsection{Results}
\begin{figure}
    \centering
    \includegraphics[width=0.7\columnwidth]{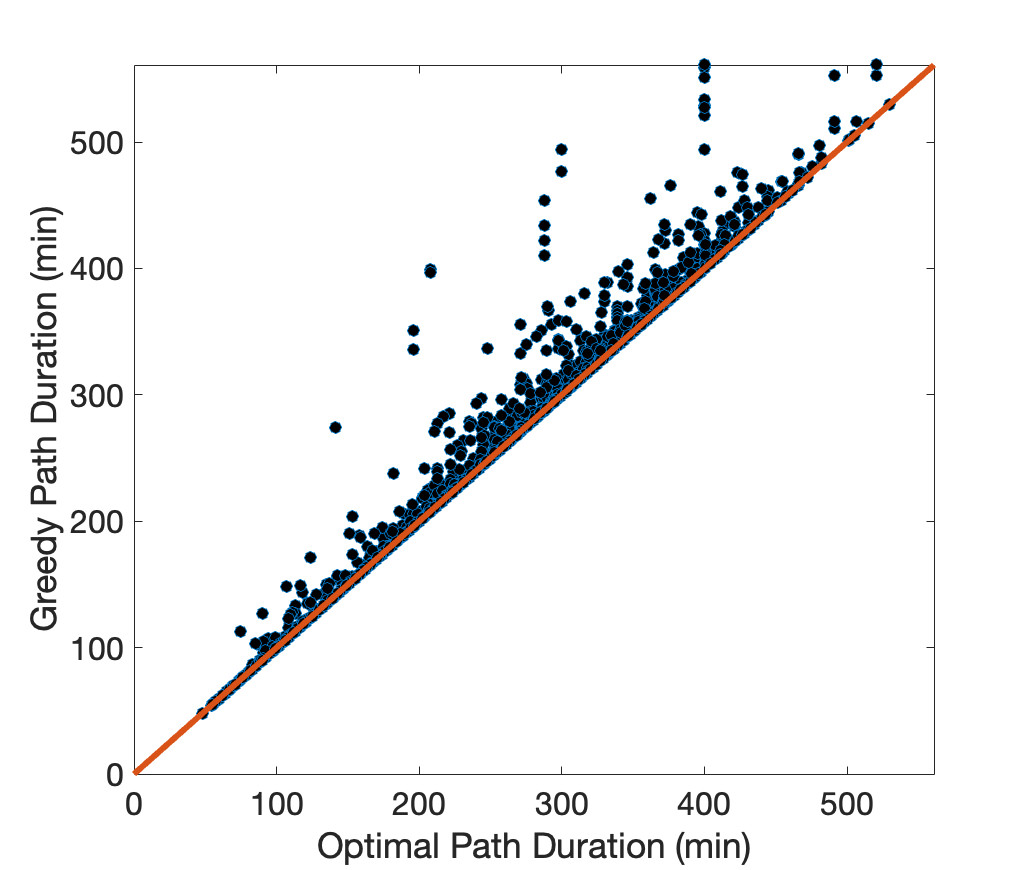}
    \caption{Performance comparison of the Greedy algorithm to the proposed Budget-$A^*$ algorithm. Each point in the graph denotes a problem instance, where the $x$ and $y$ coordinates correspond to durations of the paths generated by the Budget-$A^*$ and the Greedy algorithm respectively.  The diagonal line represents equal path duration and the vertical distance from the line indicates the difference in duration between the two solutions.\vspace{-2.0ex}}%, calculated as the difference between path times under the two algorithms divided by optimal path time.}
    \label{fig:ourFastPerformanceGrid}
\end{figure}

% Following are the simulation results comparing the performance of the proposed algorithm against the baseline methods.  We also look into some example problem instances to discuss how the greedy method can make sub-optimal decisions in generating the solution.  
Figure~\ref{fig:ourFastPerformanceGrid} compares the performance of the Budget-$A^*$ algorithm with that of the Greedy algorithm in terms of durations of the generated paths.  From the figure, we observe that the Greedy algorithm is able to generate optimal or close-to-optimal solutions for a large proportion of the tested problem instances.  However, for many instances, the path generated by the greedy approach is much longer than that produced by the Budget-$A^*$ algorithm, reaching up to twice the duration.

\begin{figure*}[ht]
    \centering
    \includegraphics[width=0.8\textwidth]{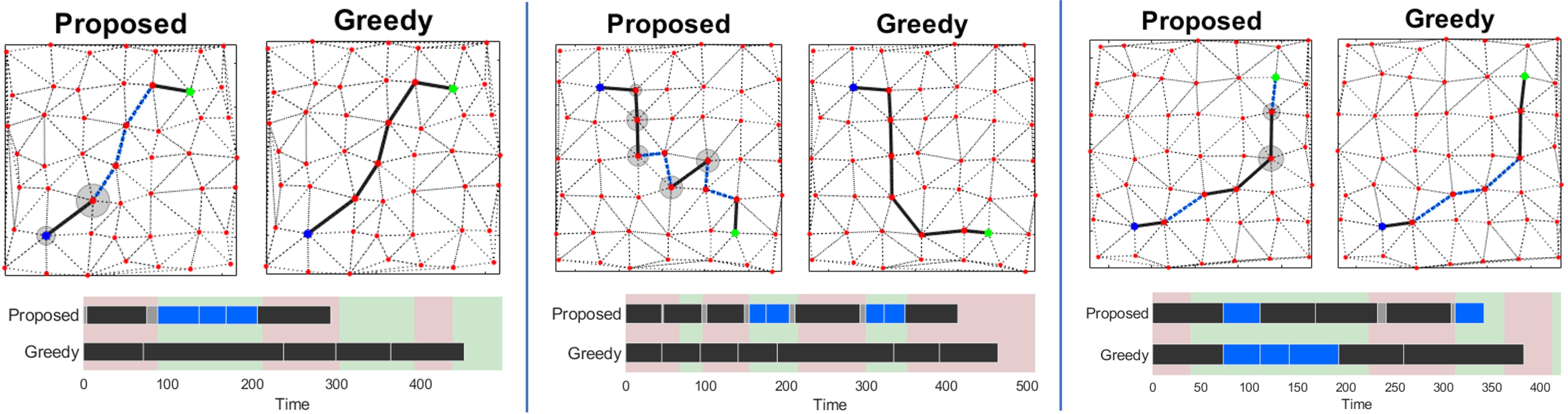}
    \caption{Example graphs comparing paths generated by the proposed Budget-$A^*$ and the Greedy algorithm. Black solid lines represent autonomous mode and blue dotted lines represent assisted mode. Grey circles denote waiting, with circle size proportional to waiting duration.  The red and greed shaded regions in the bottom plots represent operator availability.
    (a) The Budget-$A^*$ algorithm preemptively waits at initial vertices to use autonomous mode for later edges, resulting in a faster path. The greedy algorithm moves towards the goal quickly, but cannot use operator availability later due to waiting limits.
    (b) The Budget-$A^*$ algorithm uses operator availability more efficiently by selecting a longer path.
    (c) The Budget-$A^*$ algorithm chooses not to use operator's assistance even when it is available, so that it can be used later when the assistance is more beneficial.\vspace{-2.0ex}}
    \label{fig:pathComp}
\end{figure*}

To gain further insight into our results, we present Fig.~\ref{fig:pathComp}, highlighting example instances where the greedy approach fails to generate an optimal solution. Through these examples, we demonstrate how our algorithm makes effective decisions regarding path selection, preemptive waiting, and not utilizing assistance to delay arrival at a later vertex. These decisions ultimately result in improved arrival time at the goal.

\begin{figure}
    \centering
\includegraphics[width=0.87\columnwidth]{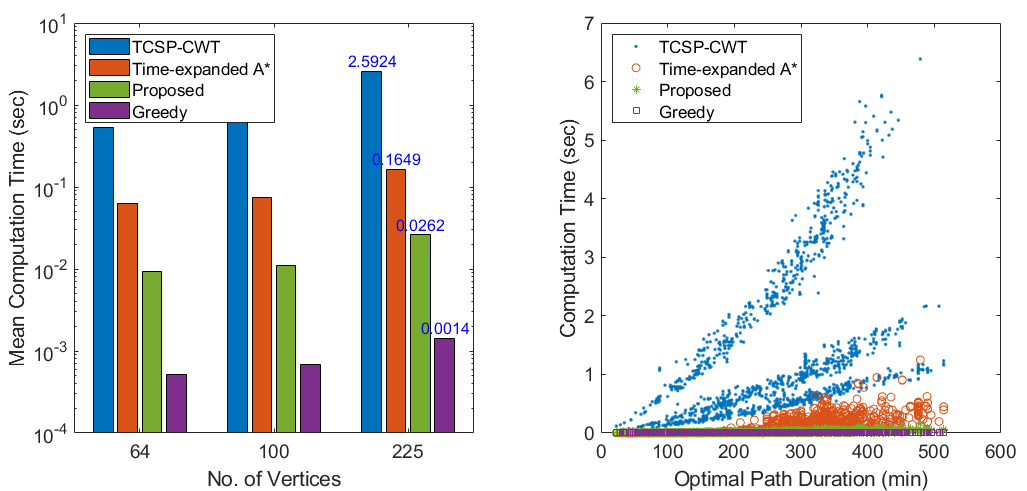}
    \caption{Computation time comparison for different methods. \textbf{Left:} Mean computation time as a function of the number of vertices in the graph, averaged over all test instances.  Note that the time is plotted on a log scale.  \textbf{Right:} Computation time as a function of actual duration of the optimal path, shown for all test instances.\vspace{-1.5ex}}
    \label{fig:timeComparison}
\end{figure}

Figure~\ref{fig:timeComparison} compares the computation time required by different solution methods for varying number of vertices and the duration of the optimal path between the start and goal vertices. The plots demonstrate that the proposed algorithm consistently outperforms the other optimal methods in terms of computation time, with the greedy method being the fastest but providing suboptimal solutions. The computation time for all methods increases with the number of vertices. The path duration has the greatest impact on the performance of the TCSP-CWT algorithm, followed by the Time-expanded $A^*$, the Budget-$A^*$ algorithm, and finally the Greedy algorithm.
% One thing to note is that the No Refinement method is faster than the No Budget method for small problems but becomes slower for larger problems.  The reason behind this is that No Refinement method uses three dimensional nodes (vertex, arrival time, budget) compared to the two dimensional nodes in No Budget method (vertex, arrival time).  While exploring nodes with a budget parameter is useful for smaller problem instances, its effectiveness diminishes with larger problems as the number of redundant nodes increases rapidly.  Thus without node refinement the budgeting method starts to fall short.
% This is observed from the performance of the Budget-$A^*$, which shows the fastest computation time among all solution methods.

% \begin{figure}
%     \centering
%     \includegraphics[width=0.8\columnwidth]{Figures/computationTimeVsEdges.png}
%     \caption{Computation time change with edges}
%     \label{fig:my_label}
% \end{figure}
% \begin{figure}
%     \centering
% \includegraphics[width=0.6\columnwidth]{Figures/computationTimevsPathTime.png}
%     \caption{Computation time comparison.}
%     \label{fig:timeComparison2}
% \end{figure}

% \subsection{Node exploration}
% In this section, we will evaluate the impact of the two key elements of our algorithm: 1) expanding nodes only at the critical times, and 2) node refinement.
% By expanding nodes at only the critical times, we can avoid unnecessary calculations and improve the efficiency of the algorithm. Additionally, the refinement process helps to ensure that the priority queue only contains the most promising nodes, further improving the speed of the algorithm. The figure compares the performance of our algorithm to the performance when these techniques are absent by comparing two measures: 1) Nodes generated and added to the priority queue, and 2) Nodes explored before finding the optimal solution.

\begin{figure}
    \centering
    \includegraphics[width=0.87\columnwidth]{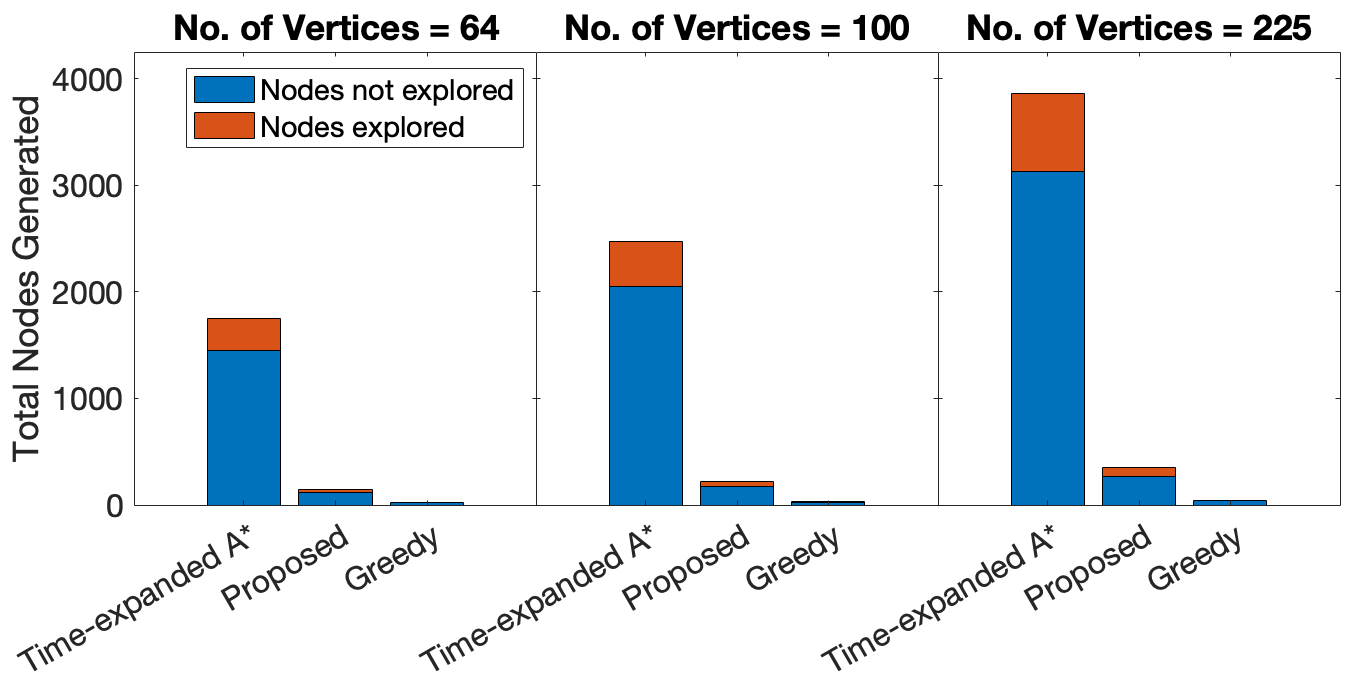}
    \caption{Mean number of nodes generated during graph search under the three methods for different number of vertices in the map.  The bar height shows the total number of nodes generated and added to the queue during the search.  The orange stack denotes the number of nodes explored before the goal is reached and the search is terminated.\vspace{-2.0ex}}
    \label{fig:nodes}
\end{figure}

Figure~\ref{fig:nodes} compares the number of nodes generated and explored by Time-expanded $A^*$, Budget $A^*$, and Greedy search algorithms.  The number of nodes is a key metric to evaluate search efficiency as it reflects the number of insertions and extractions from the priority queue.  The Time-expanded $A^*$ generates nodes at a faster rate with increasing vertices, while the proposed algorithm generates an order of magnitude fewer nodes, indicating better efficiency and scalability.  The Greedy search algorithm terminates after exploring the least number of nodes, indicating that it sacrifices optimality for speed. In contrast, both the Time-expanded $A^*$ and Budget $A^*$ algorithms guarantee optimality in their search results.

\section{Conclusion}
In this paper, we introduced Budget-$A^*$, a new algorithm to tackle the problem of collaborative robot planning with bounded waiting constraints and intermittent human availability.  Our approach computes the optimal execution path, which specifies which path should the robot take, how much to wait at each location and when to use human assistance.
Our simulations on a city road network demonstrate that Budget-$A^*$ outperforms existing optimal methods, in terms of both computation time and number of nodes explored. %The proposed algorithm efficiently finds the optimal solution even for large and complex networks.
Furthermore, we note that the greedy method performs well for the majority of test cases, which could potentially be utilized to further improve efficiency of the proposed algorithm.

For future research, the Budget-$A^*$ algorithm can be extended to handle more complex constraints such as multiple types of human assistance, non-stationary operator availability, and dynamic task requirements.  Our approach can be further optimized to handle even larger networks by incorporating better heuristics and pruning techniques.  Finally, our algorithm can be adapted to other applications such as emergency response in unknown environments, where fast and online task planning is crucial. 

Our approach has significant implications for real-world applications like transportation systems, logistics, and scheduling, where time constraints and limited human supervision are crucial. We believe our work will inspire further research in these areas and lead to the development of more efficient algorithms for enabling human supervision under real-world restrictions.

% We have also examined the impact of two key elements of our algorithm on solution speed: expanding nodes at the earliest arrival time and at times when non-FIFO transitions occur, and our refinement process which removes suboptimal or redundant nodes from the priority queue. Our results have shown that these techniques significantly improve the performance of the algorithm, leading to faster solution times and a reduction in the number of nodes explored.

%\bibliographystyle{IEEEtran}
%\bibliography{biblio}

\end{document}